\pdfoutput=1
\documentclass[11pt,a4paper]{article}
\usepackage[hyperref]{acl2019}
\usepackage{times}
\usepackage{latexsym}
\usepackage{amsmath}

\usepackage{url}
\usepackage{CJKutf8}
\usepackage{graphicx}
\usepackage{color}
\usepackage{verbatim}
\usepackage{multirow}
\usepackage{subfigure}
\usepackage{algorithm}
\usepackage{algorithmicx}
\usepackage{amsmath,amsfonts,amssymb,bbm,epsfig,bm}
\usepackage{algpseudocode}
\usepackage{balance}

\aclfinalcopy 
\def\aclpaperid{175} 

\title{Coherent Comment Generation for Chinese Articles with a Graph-to-Sequence Model}

\author{Wei Li\textsuperscript{1}, Jingjing Xu\textsuperscript{1}, Yancheng He\textsuperscript{2}, Shengli Yan\textsuperscript{2}, Yunfang Wu\textsuperscript{1}, Xu Sun\textsuperscript{1,3} \\
\textsuperscript{1}MOE Key Lab of Computational Linguistics, School of EECS, Peking University \\
\textsuperscript{2}Platform \& Content Group, Tencent \\
\textsuperscript{3}Deep Learning Lab, Beijing Institute of Big Data Research, Peking University\\
\texttt{\{liweitj47, jingjingxu\}@pku.edu.cn}\\
\texttt{\{collinhe, victoryyan\}@tencent.com} \\
\texttt{\{wuyf, xusun\}@pku.edu.cn} \\}

\date{}

\begin{document}
\maketitle
\begin{abstract}
Automatic article commenting is helpful in encouraging user engagement and interaction on online news platforms. However, the news documents are usually too long for traditional encoder-decoder based models, which often results in general and irrelevant comments. 
In this paper, we propose to generate comments with a graph-to-sequence model that models the input news as a topic interaction graph. By organizing the article into graph structure, our model can better understand the internal structure of the article and the connection between topics, which makes it better able to understand the story. 
We collect and release a large scale news-comment corpus from a popular Chinese online news platform Tencent Kuaibao.\footnote{\url{https://kuaibao.qq.com/}} Extensive experiment results show that our model can generate much more coherent and informative comments compared with several strong baseline models.\footnote{Code for the paper is available at \url{https://github.com/lancopku/Graph-to-seq-comment-generation}}
\end{abstract}

\begin{table}[th]
    \small
    \centering
    \begin{tabular}{p{6.5cm}}\hline
        \textbf{Title} \\ \hline
        \footnotesize \begin{CJK*}{UTF8}{gbsn}这部影片被称为``十年来最搞笑{\color{blue}漫威电影}''，你看了吗？\end{CJK*} \\
        Have you seen the movie intitled as ``the most hilarious {\color{blue}Marvel movie}''? \\ 
         \hline
        \textbf{Content} \\ \hline
         \footnotesize \begin{CJK*}{UTF8}{gbsn}点击``IPTV4K超高清''订阅，精彩内容等你共享《{\color{blue}复仇者联盟3}：无限战争》中的巅峰一役，将战火燃遍了整个宇宙...作为接档《{\color{blue}复联3}》的{\color{blue}漫威}电影，《{\color{blue}蚁人}2》的故事爆笑中带着温情，无疑成为了现阶段抚平漫威粉心中伤痛的一味良药...看过《{\color{blue}复联3}》的漫威粉们，心中都有同一个疑问：在几乎整个{\color{blue}复仇者联盟}都参与到无限战争的关键时刻，{\color{blue}蚁人}究竟去哪儿了？...\end{CJK*} \\
        Click on the ``IPTV4K ultra HD'' to subscribe, fantastic contents are waiting for you to share. The battle in ``{\color{blue} Avengers}: Infinity War'' has spread the flames of war throughout the universe ... As the continuation {\color{blue}Marvel} movie to ``{\color{blue}Avengers} 3'', the hilarious and warm ``{\color{blue}Ant-Man} and the Wasp'' is no doubt a good dose to heal the fans of {\color{blue}Marvel} at the time. ... Fans of the {\color{blue}Marvel} who have watched ``{\color{blue}Avengers} 3'' all have a doubt about where {\color{blue}Ant-Man} is when all other {\color{blue}Avengers} have been involved in the infinity war.\\
        \hline
        \textbf{Comment} \\ \hline
        \footnotesize \begin{CJK*}{UTF8}{gbsn}只有我觉得那个头盔像{\color{blue}蚁人}的头盔吗? \end{CJK*}\\
        Am I the only one that thinks the helmet similar to the helmet of {\color{blue}Ant-Man}?\\
         \hline
    \end{tabular}
    \caption{An example of news article comment generation task, which is to generate new comments given the title and content of the news. Because the article is too long, only the first sentence and three fragments with topic words ({\color{blue}blue}) are shown. Note that the title and the first sentence of the news are very different from traditional news, which can not summarize the content of the article.}
    \label{tab:article example}
\end{table}
\section{Introduction}

Online news platform is now a popular way for people to get information, where users also make comments or read comments made by others, making the comments very valuable resource to attract user attention and encourage interactions among users 
\citep{Park:2016:SCM:2858036.2858389}. 
The ability to automatically generate comments is desirable for online news platforms, especially comments that can encourage user engagement and interactions, serving as one form of intelligent chatbot \citep{shum2018eliza}. 
Important as the comment generation task is, it is still relatively new. \citet{qin2018automatic} proposed the problem of automatic article comment generation, which is to generate comments given the title and content of the article (An example is shown in Table \ref{tab:article example}). 
They only proposed the task, but did not propose a specially designed solution to the problem other than sequence-to-sequence paradigm \citep{sutskever2014sequence}. \citet{ma2018unsupervised} proposed a retrieval based model that uses variational topic model to find comments that are related to the news in an unsupervised fashion. \citet{lin2018learning} proposed to refer to the retrieved comments during generation, which is a combination of retrieval and generation based model. Pure generation based model remains challenging, yet is a more direct way to solve the problem. Additionally, when the article is very different from the historical ones, there may not be appropriate comments to refer to. In this work, we would like to explore a generation model that better exploits the news content to solve the problem.

Different from the scenarios where sequence-to-sequence models achieve great success like machine translation \citep{bahdanau2014neural} and summarization \citep{see2017get}, comment generation has several nontrivial challenges:
\begin{itemize}
    \item The news articles can be very long, which makes it intractable for classic sequence-to-sequence models. On the contrary, although the title is a very important information resource, it can be too short to provide sufficient information.
    \item The title of the news sometimes uses hyperbolic expressions that are semantically different from the content of the article. For example, the title shown in the example (Table \ref{tab:article example}) provides no valuable information other than ``Marvel movie'', which is far from enough to generate coherent comments.
    \item Users focus on different aspects (topics) of the news when making comments, which makes the content of the comments very diverse. For example, comments can be about the plots in ``\textit{Avengers}'', ``\textit{Ant-Man}'' or other characters in \textit{Marvel} movies.
\end{itemize}

Based on the above observations, we propose a graph-to-sequence model that generates comments based on a graph constructed out of content of the article and the title. We propose to represent the long document as a topic interaction graph, which decomposes the text into several topic centered clusters of texts, each of which representing a key aspect (topic) of the article. Each cluster together with the topic form a vertex in the graph. The edges between vertices are calculated based on the semantic relation between the vertices. Compared with the hierarchical structure \citep{yang2016hierarchical}, which is designed for long articles, our graph based model is better able to understand the connection between different topics of the news. Our model jointly models the title and the content of the article by combining the title into the graph as a special vertex, which is helpful to get the main point of the article.

We conduct extensive experiments on the news comments collected from Tencent Kuaibao news, which is a popular Chinese online news platform. We use three metrics consulting to \citet{qin2018automatic} to evaluate the generated comments. Experiment results show that our model can generate more coherent and informative  comments compared with the baseline models. 

We conclude the contributions as follows:
\begin{itemize}
    \item We propose to represent the article with a topic interaction graph, which organizes the sentences of the article into several topic centered vertices.
    \item We propose a graph-to-sequence model that generates comments based on the topic interaction graph.
    \item We collect and release a large scale (200,000) article-comment corpus that contains title, content and the comments of the news articles.
    
\end{itemize}

\section{Related Work}
The Graph Neural Networks (GNN) model has attracted growing attention recently, which is good at modeling graph structure data. GNN is not only applied in structural scenarios, where the data are naturally performed in graph structure, such as social network prediction systems \citep{NIPS2017_6703,kipf2016semi}, recommender systems \citep{van2017graph,ying2018graph}, and knowledge graphs \citep{hamaguchi2017knowledge}, but also non-structural scenarios where the relational structure is not explicit including image classification \citep{kampffmeyer2018rethinking, wang2018zero}, text, etc. In this paper, we explore to use GNN to model non-structural article text.

Some recent researches are devoted to applying GNN in the text classification task, which involves modeling long documents as graphs. \citet{peng2018large} proposed to convert a document into a word co-occurrence graph, which is then used as the input to the convolutional layers. \citet{yao2018graph} proposed to organize the words and documents into one unified graph. Edges between words are calculated with point-wise mutual information (PMI), edges between word and document are calculated with TF-IDF. Then a spectral based graph convolutional networks (GCN) is applied to classify the documents. \citet{DBLP:journals/corr/abs-1802-07459} proposed a siamese GCN model in the text matching task by modelling two documents into one interaction graph. \citet{DBLP:conf/cikm/ZhangLNLX18} adopted a similar strategy but used GCN to match the article with a short query. These works are inspiring to our work, however, they are only designed for the classification task, which are different from generation tasks.

There are also some previous work dedicated to use GNN in the generation tasks. \citet{xu2018graph2seq,xu2018sql} proposed to use graph based model to encode SQL queries in the SQL-to-Text task. \citet{beck2018graph} and \citet{song2018graph} proposed to solve the AMR-to-Text problem with graph neural networks. \citet{zhao2018graphseq2seq} proposed to facilitate neural machine translation by fusing the dependency between words into the traditional sequence-to-sequence framework. Although these work apply GNN as the encoder, they are meant to take advantage of the information that are already in the form of graph (SQL query, AMR graph, dependency graph) and the input text is relatively short, while our work tries to model long text documents as graphs, which is more challenging.

\section{Graph-to-Sequence Model}
\begin{figure*}
\centering
\includegraphics[width=6.2in]{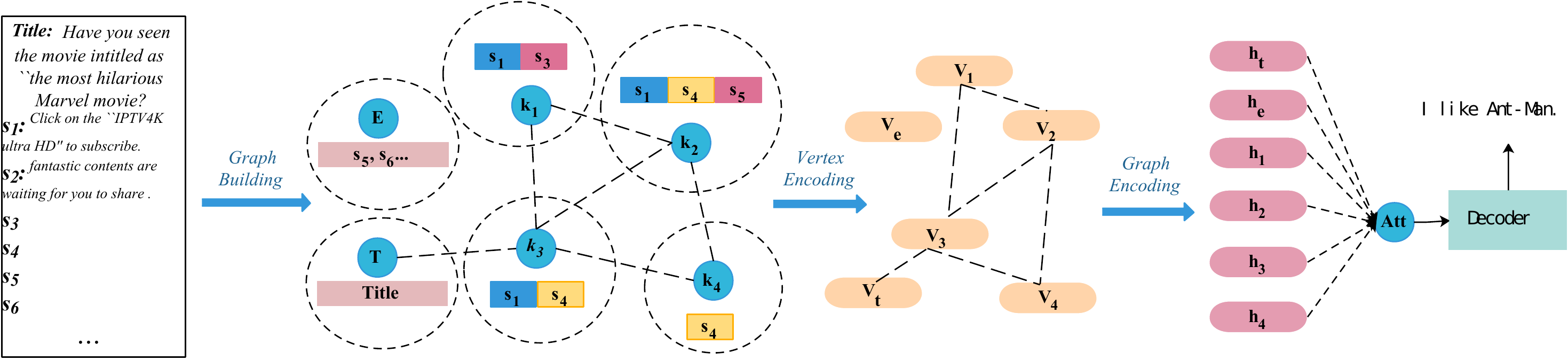}
\caption{A brief illustration of our proposed graph-to-sequence model. A vertex in the interaction graph consists of a  topic word $k_i$ and the sentences containing $k_i$. If a sentence contains no topic word, it is archived to a special ``Empty'' vertex. Each vertex is first encoded into a hidden vector $v_i$ by the vertex encoder. Then the whole graph is fed into the graph encoder and get the final vertex representation $h_i$ encoded with structure information. A RNN decoder with attention mechanism is adopted to generate comment words.}
\label{fig:graph encoder}
\end{figure*}

In this section, we introduce the proposed graph-to-sequence model (shown in Figure \ref{fig:graph encoder}). Our model follows the Encoder-Decoder framework. The encoder is bound to encode the article text presented as an interaction graph into a set of hidden vectors, based on which the decoder generates the comment sequence. 

\subsection{Graph Construction}
\label{sec:graph construction}
\setlength{\textfloatsep}{15pt}
\begin{algorithm}[t]
\centering
\footnotesize
\begin{algorithmic}[1]
\Require
The title $title$ and article text $D$, weight calculation function $\lambda$
\State
Segment $title$ and $D$ into words

\State
Do named entity recognition and keyword detection and get the keywords $\kappa$

\For{sentence $s$}
\If{$s$ contains $k\in\kappa$}
\State Assign $s$ to vertex $v_k$
\Else 
\State Assign $s$ to vertex $v_{empty}$
\EndIf
\EndFor
\For{vertex $v_i$ and $v_j$}
\State Calculate edge weight: $w_{ij} = \lambda(v_i, v_j)$
\EndFor
\end{algorithmic}
\caption{Graph Construction}
\label{alg:graph_construction}
\end{algorithm}
In this section, we introduce how to construct the topic interaction graph from a news article. Algorithm \ref{alg:graph_construction} shows the construction process.
Different from traditional news, the articles from online news platforms contain much noise. Many sentences of the articles are even irrelevant to the main topic of the news. For example,  \begin{small}\begin{CJK*}{UTF8}{gbsn}``谢谢大家点开这篇文章''\end{CJK*}\end{small} (\textit{Thanks for opening this article}). Therefore, we extract the keywords of the article which serve as the topics of the news. These keywords are the most important words to understand the story of the article, most of which are named entities. Since keyword detection is not the main point of this paper, we do not go into the details of the extraction process. 

Given a news article $D$, we first do word segmentation and named entity recognition on the news articles with off-the-shelf tools such as Stanford CoreNLP.\footnote{\url{https://stanfordnlp.github.io/CoreNLP}} Since the named entities alone can be insufficient to cover the main focuses of the document, we further apply keyword extraction algorithms like TextRank \citep{mihalcea2004textrank} to obtain additional keywords. 

After we get the keywords $\kappa$ of the news, we associate each sentence of the documents to its corresponding keywords. We adopt a simple strategy that assigns a sentence $s$ to the keyword $k$ if $k$ appears in the sentence. Note that one sentence can be associated with multiple keywords, which implicitly indicates connection between the two topics. Sentences that do not contain any of the keywords are put into a special vertex called ``\textit{Empty}''. Because the title of the article is crucial to understand the news, we also add a special vertex called ``\textit{Title}'' that contains the title sentence of the article.

The sentences together with the keyword $k$ they belong to form a vertex $v_k$ in the interaction graph. The words of the sentences are concatenated together. 
The words within each vertex represent one aspect of the article. There can be many ways to construct the edges between vertices denoted as $\lambda$ in Algorithm \ref{alg:graph_construction}. In this paper, we propose to adopt a structure based method. 
If vertices $v_i$ and $v_j$ share at least one sentence, we add an edge $e_{ij}$ between them, the weight of which is calculated by the number of shared sentences. The intuition behind this design is that the more sentences co-mention two keywords together, the closer these two keywords are. One can also use content based method such as tf-idf similarity between the content of $v_i$ and $v_j$.

\subsection{Vertex Encoder}
To encode each vertex in the graph into one hidden vector $\upsilon$, we propose to use a multi-head self-attention \citep{vaswani2017attention} based vertex encoder. 

The vertex encoder consists of two modules, the first one is an embedding module, the second one is a self-attention module. For the $i$-th word $w_i$ in the word sequence, we first look up the word embedding of the words $e_i$. Note that the keywords and regular words in the article share the same embedding table. By ``regular words'' we mean words other than keywords. To represent the position information of each word, a positional embedding $p_i$ is added to the word. The keyword $k$ of the vertex is put in the front of the word sequence. Therefore, the positional embedding of all the inserted keywords share the same embedding $p_0$, which indicates the special role of the keyword. Both the word embedding and positional embedding are set to be learn-able vectors. The final embedding $\epsilon_i$ of word $w_i$ is the sum of the original word embedding $e_i$ and positional embedding $p_i$,
$$\epsilon_i = e_i + p_i$$
Then we feed $\epsilon_i$ to the self-attention module and get the hidden vector $a_i$ of each word. This module is to model the interaction between the words so that each hidden vector in this layer contains the context information of the vertex. The self-attention module contains multiple layers of multi-head self-attention. The hidden vector of each layer is calculated by Equation (\ref{eqn: self-attention 1})-(\ref{eqn: self-attention 3}), where $Q, K, V$ represent query vector, key vector and value vectors respectively. In our case, $Q, K, V$ all represent the same vectors. For the first layer, they are $\epsilon$. For the following layers, they are the hidden vectors calculated by the previous layer. $W^o, W_i^Q, W_i^K, W_i^V$ are all learnable matrices,

\vspace{-2mm}
\begin{small}
    \begin{align}
         Attention(Q,K,V)  = & softmax(QK^T)V \label{eqn: self-attention 1}  \\ 
        MultiHead(Q,K,V)  = &
        [head_1; \cdots; head_h]W^o \label{eqn: self-attention 2} \\ 
        head_i  = Attention (&QW_i^Q, KW_i^K, VW_i^V)  \label{eqn: self-attention 3} 
    \end{align}
\end{small}
Since the keyword $k$ is the most important information in the vertex, we use the hidden vector of the inserted keyword $a_0$ in the last layer as the vector that represents the whole vertex.

\subsection{Graph Encoder}
After we get the hidden vector of each vertex $v_i$ in the graph, we feed them to a graph encoder to make use of the graph structure of the constructed topic interaction graph. We propose to use spectral based graph convolutional model (GCN). Spectral approaches work with a spectral representation
of the graphs \citep{DBLP:journals/corr/abs-1812-08434}. We choose this architecture because GCN can both model the content of the vertex and make use of the structure information of the graph. 

We use an implementation of GCN model similar to the work of \citet{kipf2016semi}. Denote the adjacency matrix of the interaction graph as $A\in R^{N\times N}$, where $A_{ij} = w_{ij}$ (defined in Section \ref{sec:graph construction}). We add an edge that points to the node itself (Equation \ref{eqn:self_adj}). $D$ is a diagonal matrix where $\tilde{D}_{ii} = \sum_j\tilde{A}_{ij}$, 
\begin{eqnarray}
    H^{l+1} = \sigma(\tilde{D}^{-\frac{1}{2}}\tilde{A}\tilde{D}^{-\frac{1}{2}}H^l W^l)\\
    \tilde{A} = A + I_N \label{eqn:self_adj} 
\end{eqnarray}
where $I_N$ is the identity matrix, $\tilde{D}^{-\frac{1}{2}}\tilde{A}\tilde{D}$ is the normalized symmetric adjacency matrix, $W^l$ is a learnable weight matrix. To avoid the over-smoothing problem of GCN, we add residual connections between layers,
\begin{eqnarray}
    g^{l+1} = H^{l+1} + H^{l} \\
    g^{out} = tanh(W_o g^K)
\end{eqnarray}
We add one feed forward layer to the final output of the GCN. $g^K$ is the output of the last layer of GCN.

Since the title of the news is still an important information, we use the hidden output of the title vertex of the graph encoder as the initial state $t_0$ of the decoder. One can also use other pooling method such as max pooling or mean pooling.



\subsection{Decoder}
For the decoder, we adopt the recurrent neural network (RNN) decoder with attention mechanism \citep{bahdanau2014neural}. Given the initial state $t_0$ and the output of the GCN $\left \langle g_0, g_1, \cdots, g_n \right \rangle$, the decoder is bound to generate a sequence of comment tokens $y_1, y_2, \cdots, y_m$. At each decoding step, a context vector $c_i$ is calculated by doing attention on the outputs of the GCN, 
\begin{eqnarray}
    t_i = RNN(t_{i-1}, e_{i-1}) \\
    c_i = \sum\alpha_j\times g_j \\
    \alpha_j = \frac{exp(\delta(t_i, g_j)}{\sum exp(\delta(t_i, g_k))}
\end{eqnarray}
where $\delta$ is the attention function.

Since the topic words (name of the vertices) $\kappa$ are important information for the article and may appear in the comment, we adopt copy mechanism \citep{DBLP:journals/corr/GuLLL16} by merging the predicted word token probability distribution with the attention distribution. The probability $p_{copy}$ of copying  from the topic words is dynamically calculated with the decoding hidden state $t_i$ and the context vector $c_i$,

\vspace{-2mm}
\begin{small}
    \begin{eqnarray}
         y_i = softmax(W_o(tanh(W([t_i; c_i])+b)))  \\
         p_{copy} = \sigma(W_{copy} [t_i; c_i])\\
         p = (1-p_{copy})\times y + p_{copy}\times \alpha
    \end{eqnarray}
\end{small}
where $W_o, W, W_{copy}, b$ are all learnable parameters.

\section{Experiment}
\subsection{Corpus}
We collect news and comments from Tencent Kuaibao,\footnote{\url{https://kuaibao.qq.com/}} which is a popular online news platform in Chinese. Because the number of news is very large and the comments vary a lot between different topics of news, we select the news from two most popular topics (topics that have the most news and comments) \textit{Entertainment} and \textit{Sport}. The data is available at \url{https://pan.baidu.com/s/1b5zAe7qqUBmuHz6nTU95UA}\footnote{The extraction code is 6xdw}. The document number and comment number of the two topics are listed in Table \ref{tab:document number}. 

\begin{table}[t]
    \centering
    \begin{tabular}{c|c|c}\hline
        Topic &  document \# & comment \# \\ \hline
        Entertainment & 116,138 & 287,889 \\
        Sport & 90,979 & 378,677 \\ \hline
    \end{tabular}
    \caption{Document and comment number of Entertainment and Sport.}
    \label{tab:document number}
\end{table}

The average length with respect to words and characters of content, title, comment and keyword for the two topics are listed in Table \ref{tab:corpus stats}. From the number we can see that the length of news content is too large for traditional sequence-to-sequence model. 

\begin{table}[t]
    \centering
    \begin{tabular}{|c|c|c|c|c|} \hline
     \multirow{2}{*}{}& \multicolumn{2}{c|}{average word \#} & \multicolumn{2}{c|}{average character \#} \\ \cline{2-5}
     & Ent & Sport & Ent & Sport \\ \hline
     content  &  456.1 & 506.6 & 754.0 & 858.7\\
     title    &  16.4 & 15.7 & 28.1 & 27.4\\
     comment  &  16.3 & 19.4 & 26.2 & 31.2\\
     keyword  & 8.4 & 9.0 & - & - \\ \hline
    \end{tabular}
    \caption{Length of content, title, comment and keyword of the news for the topic of \textbf{Ent} (entertainment) and \textbf{Sport}.}
    \label{tab:corpus stats}
\end{table}

\subsection{Experiment Settings} \label{sec: settings}
We use a batch size of 32. The embedding size is set to 128. The word embeddings are shared between encoder and decoder. Because the vertex number (keyword number in Table \ref{tab:corpus stats}) is relatively small, to ease the over-smoothing problem we use 1-layer convolution in GCN. For all the RNN based encoders, we use bidirectional LSTM and set the hidden size to 128. For the baseline hierarchical attention model, the hidden size of the second LSTM layer is 256. We use a  vocabulary size of 60,000. The sentences are truncated to 100 words. The maximum length for generating is set to 32. For multi-head attention, we use 4 heads. For RNN encoder, RNN decoder and multi-layer self-attention, we use a layer number of 2. We use a dropout rate of 0.1. We use Adam optimizer \citep{kingma2014adam} to train the parameters. The initial learning rate is set to 0.0005. For all the models, we train for 5 epochs, the learning rate is decayed to half after each epoch.

\subsection{Evaluation Metrics}
We choose three metrics to evaluate the quality of generated comments. For all the metrics, we ask the raters to score the comments with three gears, the scores are then projected to $0 \sim 10$.
\begin{itemize}
    \item \textbf{Coherence}: This metric evaluates how Coherent (consistent) is the comment to the news document. It measures whether the comment is about the main story of the news, one side part of the news, or irrelevant to the news.
    \item \textbf{Informativeness}: This metric evaluates how much concrete information the comment contains. It measures whether the comment involves a specific aspect of some character or event, or is a general description of some character or event, or is a general comment that can be the answer to many news.
    \item \textbf{Fluency}: This metric evaluates whether the sentence is fluent. It mainly measures whether the sentence follows the grammar and whether the sentence accords with the logic including world knowledge.
\end{itemize}

We ask three raters to evaluate the generated comments of different models. Owing to the laborious evaluation process (reading the long news document is time consuming), we ask the raters to evaluate the generated comments from one hundred news documents of both topics. The raters are given both the title and the document content of the news which is the same as how a user would read the news online.

We use spearman's rank score to measure the correlation between raters.
The p-values are all below $1e-50$. The ratings between raters have relatively good correlation with spearman's rank of around 0.6. Among the metrics, fluency is more divergent. This is expected as this metric is more flexible,  different people may have more divided opinion.


\begin{table*}[t]
    \centering
    \begin{tabular}{|c|c|c|c|c|} \hline
    Models & Coherence & Informativeness & Fluency & Total \\ \hline
    seq2seq-T \citep{qin2018automatic} & 5.38 & 3.70 & 8.22 & 5.77  \\
    seq2seq-C \citep{qin2018automatic} & 4.87 & 3.72 & 8.53 & 5.71 \\
    seq2seq-TC \citep{qin2018automatic} & 3.28 & 4.02 & 8.68 & 5.33 \\
    \hline
    self-attention-B \citep{DBLP:journals/corr/abs-1804-09849} & 6.72 & 5.05 & 8.27 & 6.68 \\
    self-attention-K \citep{DBLP:journals/corr/abs-1804-09849} & 6.62 & 4.73 & 8.28 & 6.54 \\ \hline
    hierarchical-attention \citep{yang2016hierarchical} & 1.38 & 2.97 & \textbf{8.65} & 4.33 \\ \hline
    graph2seq (proposed) & \textbf{8.23} & \textbf{5.27} & 8.08 & \textbf{7.19} \\ \hline
    \end{tabular}
    \caption{Comparison between our graph2seq model and baseline models for the topic of \textbf{entertainment}. T, C, B, K represents title, content, bag of words, keywords separately. \textit{Total} is the average of other three metrics}
    \label{tab:comparison entertainment}
\end{table*}
\begin{table*}[t]
    \centering
    \begin{tabular}{|c|c|c|c|c|} \hline
    Models & Coherence & Informativeness & Fluency & Total \\ \hline
    seq2seq-T \citep{qin2018automatic} & 4.30 & 4.38 & 6.27 & 4.98  \\
    seq2seq-C \citep{qin2018automatic} & 3.88 & 3.85 & 6.02 & 4.58 \\
    seq2seq-TC \citep{qin2018automatic} & 4.70 & 5.08 & 6.37 & 5.38  \\
    \hline
    self-attention-B \citep{DBLP:journals/corr/abs-1804-09849} & 5.15 & 5.62 & 6.28 & 5.68 \\
    self-attention-K \citep{DBLP:journals/corr/abs-1804-09849} & 6.68 & 5.83 & \textbf{7.00} & 6.50 \\ \hline
    hierarchical-attention \citep{yang2016hierarchical} & 4.43 & 5.05 & 6.02 & 5.17 \\ \hline
    graph2seq (proposed) & \textbf{7.97} & \textbf{6.18} & 6.37 & \textbf{6.84} \\ \hline
    \end{tabular}
    \caption{Comparison between our graph2seq model and baseline models for the topic of \textbf{sport}. T, C, B, K represents title, content, bag of words, keywords separately. \textit{Total} is the average of other three metrics}
    \label{tab:comparison sport}
\end{table*}

\subsection{Baseline Models}
In this section, we describe the baseline models we use. The settings of these models are described in Section \ref{sec: settings}. Note that for fair comparison, all the baselines use RNN with attention as the decoder, the choice of the encoder is dependent on the input of the model (whether the input is in order or not).
\begin{itemize}
    \item \textbf{Seq2seq} \citep{qin2018automatic}: this model follows the framework of sequence-to-sequence model with attention. We use three kinds of input, the title (\textbf{T}), the content (\textbf{C}) and the title together with the content (\textbf{TC}). The length of the input sequence is truncated to 100. For the input of title together with content, we append the content to the back of the title.
    \item \textbf{Self-attention} \citep{DBLP:journals/corr/abs-1804-09849}: this model follows the encoder-decoder framework. We use multi-layer self-attention with multi-head as the encoder, and a RNN decoder with attention is applied. We use two kinds of input, the bag of words (\textbf{B}) and the keywords (\textbf{K}). Since the input is not sequential, positional encoding is not applied. A special `\textit{CLS}' label is inserted, the hidden vector of which serves as the initial state of decoder.  For the bag of words input we use the words with top 100 term frequency (TF) in the news document. For the keywords input, we use the same extracted keywords (topic words) with the ones used in our topic interaction graph.
    \item \textbf{Hierarchical-Attention} \citep{yang2016hierarchical}: this model takes all the content sentences as input and applies hierarchical attention as the encoder to get the sentence vectors and document vector. A RNN decoder with attention is applied. The document vector is used as the initial state for RNN decoder.
\end{itemize}

\subsection{Results}

In Table \ref{tab:comparison entertainment} and Table \ref{tab:comparison sport}, we show the results of different baseline models and our graph2seq model for the topic of entertainment and sport separately. From the results we can see that our proposed graph2seq model beats all the baselines in both coherence and informativeness.

\noindent \textbf{Coherence}:
Our model receives much higher scores in coherence compared with all other baseline models. This indicates that our graph based model can better get the main point of the article instead of referring to the high frequency terms that are only slightly related or even irrelevant to the article, which is often carried out by baseline models (especially seq2seq based models). Besides, other baseline models tend to generate general comments such as ``\textit{I still think I like him}'' when encountering low frequency topics (similar to the dull response problem in dialogue). These two phenomena hurt the coherence performance severely. Compared with other baselines, self-attention based models receive higher coherence score, we assume that this is because the most relevant words are maintained by the bag of words and keywords input. However, it is hard to distinguish the main point of the article from all other    input words with self-attention model. Therefore, they do not perform as well as our graph based model, which can make use of the structure of the article. For the hierarchical attention model, although it uses a hierarchical structure to organize the article, it is still very difficult for the model to understand the story. In fact, we observe in the experiment that the hierarchical structure even makes it harder to extract useful information because of the oversimplified attention performed in the word level. 

\begin{table*}[t]
    \centering
    \small
    \begin{tabular}{|p{1.6cm}|p{13cm}|}\hline
        Title &  \begin{CJK*}{UTF8}{gbsn}被王丽坤美到了，《上新了·故宫》里穿古装温婉又娴静，气质惊艳\end{CJK*} $\qquad\qquad\qquad\qquad\qquad$ \textit{In ``updates of the Palace Museum'' Likun Wang appears so gentle, refined and astonishingly elegant wearing ancient costume that audiences are touched by her beauty.} \\ \hline
         S2S-T & \begin{CJK*}{UTF8}{gbsn}我觉得还是喜欢看的古装，古装扮相，古装扮相很好看\end{CJK*}$\qquad\qquad\qquad$\textit{I still think I like ancient costume, appearance in ancient costume, appearance in ancient costume is pretty.} \\ \hline
         S2S-C & \begin{CJK*}{UTF8}{gbsn}我觉得还是喜欢看的\end{CJK*}$\qquad\qquad\qquad\qquad\qquad$\textit{I still think I like to watch}\\ \hline
         S2S-TC & \begin{CJK*}{UTF8}{gbsn}我觉得还是喜欢看的\end{CJK*}$\qquad\qquad\qquad\qquad\qquad$\textit{I still think I like to watch}\\ \hline
         SA-B & \begin{CJK*}{UTF8}{gbsn}我觉得赵丽颖的演技真的很好\end{CJK*}$\qquad\qquad\qquad$\textit{I think the acting skill of Liying Zhao is very good }\\ \hline
         SA-K & \begin{CJK*}{UTF8}{gbsn}我觉得还是喜欢李沁\end{CJK*}$\qquad\qquad\qquad\qquad\qquad$\textit{I still think I like Qin Li}\\ \hline
         HA & \begin{CJK*}{UTF8}{gbsn}我觉得还是喜欢看她的剧\end{CJK*}$\qquad\qquad\qquad\qquad$\textit{I still think I like her plays}\\ \hline
         graph2seq & \begin{CJK*}{UTF8}{gbsn}王丽坤的演技真的好\end{CJK*}$\qquad\qquad\qquad\qquad\qquad$\textit{The acting skill of Likun Wang is really good} \\ \hline
    \end{tabular}
    \caption{An example of comments generated by different models. \textbf{Title} is the original title of the article. \textit{S2S}, \textit{SA}, \textit{HA} indicate seq2seq, self-attention and hierarchical attention respectively. T, C, B, K represents title, content, bag of words, keywords separately.}
    \label{tab:case study}
\end{table*}

\begin{figure}
    \centering
    \hspace{-5mm}
    \subfigure{
    \begin{minipage}[t]{0.5\linewidth}
    \centering
    \includegraphics[width=1.65in]{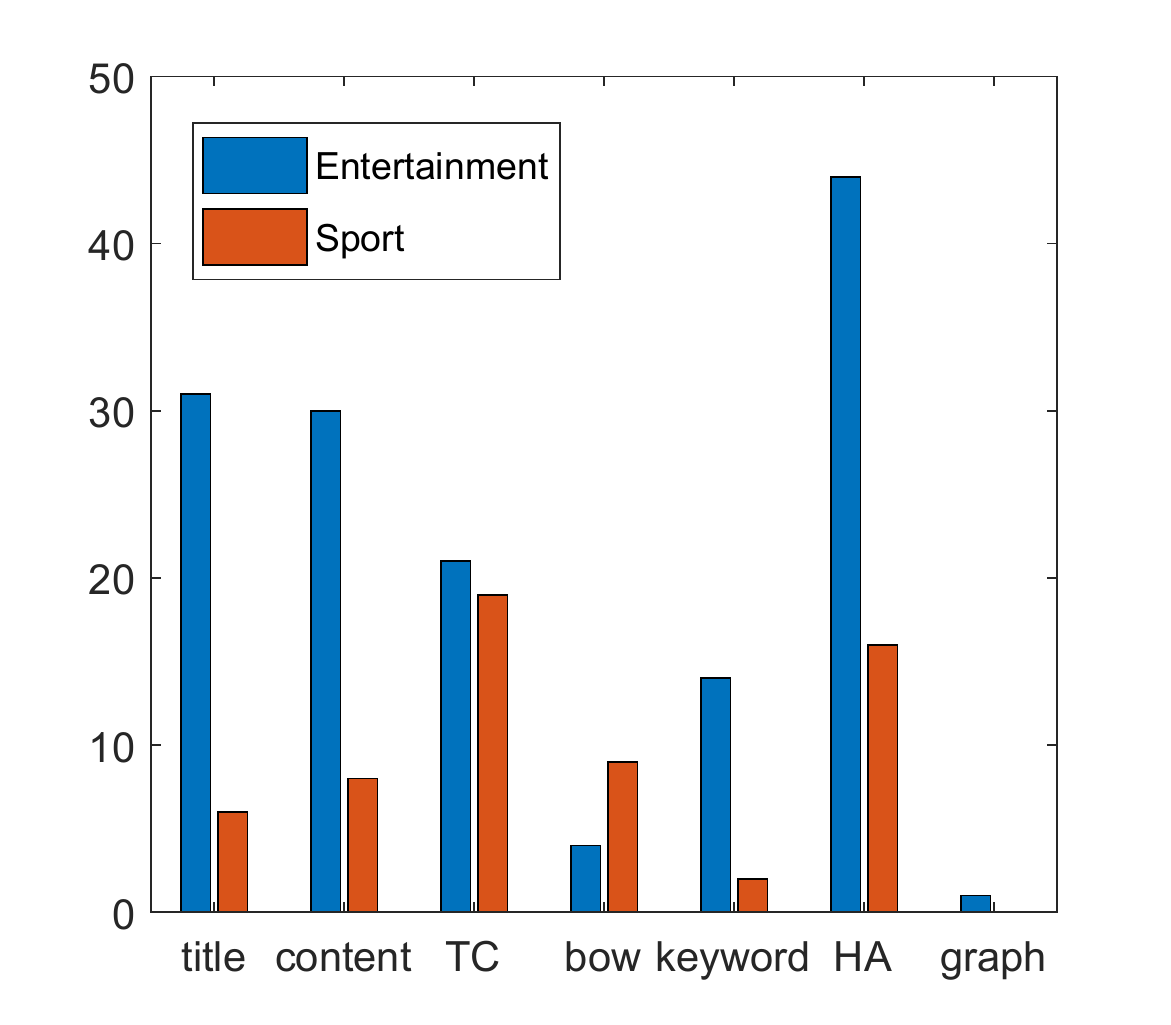}
    \end{minipage}%
    }%
    \subfigure{
    \begin{minipage}[t]{0.5\linewidth}
    \centering
    \includegraphics[width=1.65in]{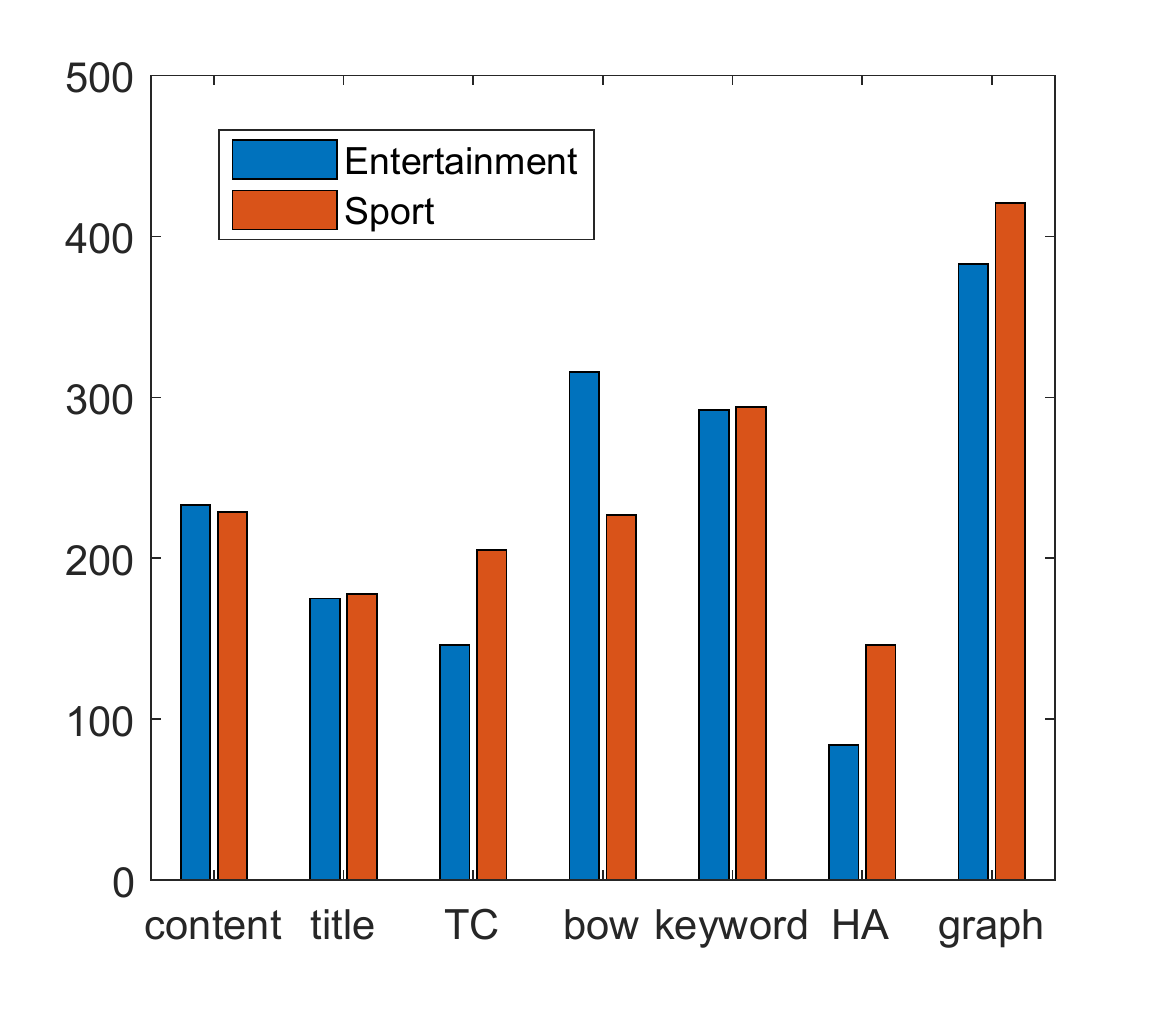}
    \end{minipage}%
    }%
    \caption{Number of generated \textbf{general comments} (Left, the lower the better) and number of \textbf{unique words} (Right, the higher the better) in the generated comments by different models. The comments from a total number of 100 news articles are inspected.}
    \label{fig:general comment}
\end{figure}

\noindent \textbf{Informativeness}:
For the metric of informativeness, our graph2seq model can generate comments with the most information because it can capture the plot of the article. We observe that this metric is related to the metric of coherence. Models with higher coherence score tend to be more informative. This phenomenon is related to the fact that many of the comments with low informative scores are general comments which are naturally not coherent to the news. In Figure \ref{fig:general comment} we show the number of generated \textit{general} comments and number of generated \textit{unique} words for both topics. By ``\textit{general comment}'', we mean those comments that have no specific information, irrelevant to the news and can be the comment to many other news of different stories, e.g., ``\textit{I still think I like him}''. Note that the notion of \textit{general comment} is not strictly defined, but an information that is meant to help analyze \textit{informativeness} score. The unique words are those not in a pre-defined stop word list. From the figure we can see that the number of general comments is loosely negatively correlated to the informative score, especially in entertainment topic. The number of generated unique words can also be an indicator for the informativeness of the comments, because the more words are involved in the comment, the more information the comment is able to provide.  


\noindent \textbf{Fluency}:
Our model receives comparable fluency score in the experiments, we assume that this is because of the similar structure of decoder between different models. After inspecting a part of the generated comments, we observe that the following reasons may lead to low fluency cases.
                         
(1) The generated comment is against the world knowledge, for instance, ``The big feast is a good actor (\begin{small}
\begin{CJK*}{UTF8}{gbsn}大餐是个好演员\end{CJK*}
\end{small})''. 

(2) The model can not distinguish between similar characters, for instance, ``\textit{Who is Han Lu? I only know Han Lu} \begin{small}
\begin{CJK*}{UTF8}{gbsn}(鹿晗是谁？我只认识鹿晗)\end{CJK*}
\end{small}''. 

(3) The model sometimes repeatedly generates the same names. We assume that this is because repeated pattern appears in some of the real comments and the copy mechanism sometimes makes the problem more severe.

These phenomena are actually observed in comments generated by various models, problems such as the deficiency of understanding world knowledge are actually very hard to solve, which are beyond the discussion of this paper.

\subsection{Case Study}

In Table \ref{tab:case study} we show an example of comments generated by different models. 

For the seq2seq-\textbf{T} (\textbf{S2S-T}) model \citep{qin2018automatic}, the comment is generated mainly based on the clue ``ancient costume'' in the title. However, because ``ancient costume'' is not frequently seen in the comments (in the training set). The pattern of generating comments about ``ancient costume'' is not well learned by the model, which makes the language of the comment not fluent.
The comment generated by the seq2seq-\textbf{C} (\textbf{S2S-C}) model is a typical \textit{general comment}, which includes no specific information. This happens when the input to the model does not contain obvious signals that indicates what topic the comment should be about. Despite the fact that these comments are not what we desire, these comments get good fluency scores, which explains why the fluency scores of some of the baselines exceed our model's. 
The comment made by hierarchical attention model (\textbf{HA}) suffers from the same problem with seq2seq model. We assume that this is because even with the hierarchical structure, this model can not understand the long input well. Therefore, it can not extract the main point of the story and generate general comments.

The comments made by self-attention based models (\textbf{SA}) are generally more informative, which contain more specific plots or characters. Even though the input to these models are not in order, the combination of the keywords makes the model easier to associate the input with some learned pattern. However, this way of representing the article is incapable of getting the main point of the article. The main characters in the generated comments ``\begin{small}
\begin{CJK*}{UTF8}{gbsn}赵丽颖\end{CJK*}
\end{small}'' and ``\begin{small}
\begin{CJK*}{UTF8}{gbsn}李沁\end{CJK*}
\end{small}'' (names of Chinese actresses)  are not much related to the news.

The comment generated by our proposed graph2seq model is the only model that mentions the main character of the news \begin{small}
\begin{CJK*}{UTF8}{gbsn}``王丽坤''\end{CJK*}
\end{small} (name of the Chinese actress), which accords with the expectation of the design of our graph based model.

\section{Conclusion}
In this paper, we propose to automatically generate comment of articles with a graph-to-sequence model that organizes the article into a topic interaction graph. Our model can better understand the structure of the article, thus capturing the main point of the article. Experiment results show that our model can generate more coherent and informative comments. We observe that there are still some comments conflicting with the world knowledge. In the future, we would like to explore how to introduce external knowledge into the graph to make the generated comments more logical.

\section*{Acknowledgement}
We thank the anonymous reviewers for their thoughtful comments. This work was supported in part by National Natural Science Foundation of China (No. 61673028). Xu Sun is the corresponding author of this paper.

\bibliography{acl2019}

\begin{thebibliography}{31}
\expandafter\ifx\csname natexlab\endcsname\relax\def\natexlab#1{#1}\fi

\bibitem[{Bahdanau et~al.(2014)Bahdanau, Cho, and Bengio}]{bahdanau2014neural}
Dzmitry Bahdanau, Kyunghyun Cho, and Yoshua Bengio. 2014.
\newblock Neural machine translation by jointly learning to align and
  translate.
\newblock \emph{arXiv preprint arXiv:1409.0473}.

\bibitem[{Beck et~al.(2018)Beck, Haffari, and Cohn}]{beck2018graph}
Daniel Beck, Gholamreza Haffari, and Trevor Cohn. 2018.
\newblock Graph-to-sequence learning using gated graph neural networks.
\newblock \emph{arXiv preprint arXiv:1806.09835}.

\bibitem[{van~den Berg et~al.(2017)van~den Berg, Kipf, and
  Welling}]{van2017graph}
Rianne van~den Berg, Thomas~N Kipf, and Max Welling. 2017.
\newblock Graph convolutional matrix completion.
\newblock \emph{stat}, 1050:7.

\bibitem[{Chen et~al.(2018)Chen, Firat, Bapna, Johnson, Macherey, Foster,
  Jones, Parmar, Schuster, Chen, Wu, and
  Hughes}]{DBLP:journals/corr/abs-1804-09849}
Mia~Xu Chen, Orhan Firat, Ankur Bapna, Melvin Johnson, Wolfgang Macherey,
  George Foster, Llion Jones, Niki Parmar, Mike Schuster, Zhifeng Chen, Yonghui
  Wu, and Macduff Hughes. 2018.
\newblock \href {http://arxiv.org/abs/1804.09849} {The best of both worlds:
  Combining recent advances in neural machine translation}.
\newblock \emph{CoRR}, abs/1804.09849.

\bibitem[{Gu et~al.(2016)Gu, Lu, Li, and Li}]{DBLP:journals/corr/GuLLL16}
Jiatao Gu, Zhengdong Lu, Hang Li, and Victor O.~K. Li. 2016.
\newblock \href {http://arxiv.org/abs/1603.06393} {Incorporating copying
  mechanism in sequence-to-sequence learning}.
\newblock \emph{CoRR}, abs/1603.06393.

\bibitem[{Hamaguchi et~al.(2017)Hamaguchi, Oiwa, Shimbo, and
  Matsumoto}]{hamaguchi2017knowledge}
Takuo Hamaguchi, Hidekazu Oiwa, Masashi Shimbo, and Yuji Matsumoto. 2017.
\newblock Knowledge transfer for out-of-knowledge-base entities: a graph neural
  network approach.
\newblock \emph{arXiv preprint arXiv:1706.05674}.

\bibitem[{Hamilton et~al.(2017)Hamilton, Ying, and Leskovec}]{NIPS2017_6703}
Will Hamilton, Zhitao Ying, and Jure Leskovec. 2017.
\newblock \href
  {http://papers.nips.cc/paper/6703-inductive-representation-learning-on-large-graphs.pdf}
  {Inductive representation learning on large graphs}.
\newblock In I.~Guyon, U.~V. Luxburg, S.~Bengio, H.~Wallach, R.~Fergus,
  S.~Vishwanathan, and R.~Garnett, editors, \emph{Advances in Neural
  Information Processing Systems 30}, pages 1024--1034. Curran Associates, Inc.

\bibitem[{Kampffmeyer et~al.(2018)Kampffmeyer, Chen, Liang, Wang, Zhang, and
  Xing}]{kampffmeyer2018rethinking}
Michael Kampffmeyer, Yinbo Chen, Xiaodan Liang, Hao Wang, Yujia Zhang, and
  Eric~P Xing. 2018.
\newblock Rethinking knowledge graph propagation for zero-shot learning.
\newblock \emph{arXiv preprint arXiv:1805.11724}.

\bibitem[{Kingma and Ba(2014)}]{kingma2014adam}
Diederik~P Kingma and Jimmy Ba. 2014.
\newblock Adam: A method for stochastic optimization.
\newblock \emph{arXiv preprint arXiv:1412.6980}.

\bibitem[{Kipf and Welling(2016)}]{kipf2016semi}
Thomas~N Kipf and Max Welling. 2016.
\newblock Semi-supervised classification with graph convolutional networks.
\newblock \emph{arXiv preprint arXiv:1609.02907}.

\bibitem[{Lin et~al.(2018)Lin, Winata, and Fung}]{lin2018learning}
Zhaojiang Lin, Genta~Indra Winata, and Pascale Fung. 2018.
\newblock Learning comment generation by leveraging user-generated data.
\newblock \emph{arXiv preprint arXiv:1810.12264}.

\bibitem[{Liu et~al.(2018)Liu, Zhang, Niu, Lin, Lai, and
  Xu}]{DBLP:journals/corr/abs-1802-07459}
Bang Liu, Ting Zhang, Di~Niu, Jinghong Lin, Kunfeng Lai, and Yu~Xu. 2018.
\newblock \href {http://arxiv.org/abs/1802.07459} {Matching long text documents
  via graph convolutional networks}.
\newblock \emph{CoRR}, abs/1802.07459.

\bibitem[{Ma et~al.(2018)Ma, Cui, Wei, and Sun}]{ma2018unsupervised}
Shuming Ma, Lei Cui, Furu Wei, and Xu~Sun. 2018.
\newblock Unsupervised machine commenting with neural variational topic model.
\newblock \emph{arXiv preprint arXiv:1809.04960}.

\bibitem[{Mihalcea and Tarau(2004)}]{mihalcea2004textrank}
Rada Mihalcea and Paul Tarau. 2004.
\newblock Textrank: Bringing order into text.
\newblock In \emph{Proceedings of the 2004 conference on empirical methods in
  natural language processing}.

\bibitem[{Park et~al.(2016)Park, Sachar, Diakopoulos, and
  Elmqvist}]{Park:2016:SCM:2858036.2858389}
Deokgun Park, Simranjit Sachar, Nicholas Diakopoulos, and Niklas Elmqvist.
  2016.
\newblock \href {https://doi.org/10.1145/2858036.2858389} {Supporting comment
  moderators in identifying high quality online news comments}.
\newblock In \emph{Proceedings of the 2016 CHI Conference on Human Factors in
  Computing Systems}, CHI '16, pages 1114--1125, New York, NY, USA. ACM.

\bibitem[{Peng et~al.(2018)Peng, Li, He, Liu, Bao, Wang, Song, and
  Yang}]{peng2018large}
Hao Peng, Jianxin Li, Yu~He, Yaopeng Liu, Mengjiao Bao, Lihong Wang, Yangqiu
  Song, and Qiang Yang. 2018.
\newblock Large-scale hierarchical text classification with recursively
  regularized deep graph-cnn.
\newblock In \emph{Proceedings of the 2018 World Wide Web Conference on World
  Wide Web}, pages 1063--1072. International World Wide Web Conferences
  Steering Committee.

\bibitem[{Qin et~al.(2018)Qin, Liu, Bi, Wang, Liu, Hu, Zhao, and
  Shi}]{qin2018automatic}
Lianhui Qin, Lemao Liu, Wei Bi, Yan Wang, Xiaojiang Liu, Zhiting Hu, Hai Zhao,
  and Shuming Shi. 2018.
\newblock Automatic article commenting: the task and dataset.
\newblock \emph{arXiv preprint arXiv:1805.03668}.

\bibitem[{See et~al.(2017)See, Liu, and Manning}]{see2017get}
Abigail See, Peter~J Liu, and Christopher~D Manning. 2017.
\newblock Get to the point: Summarization with pointer-generator networks.
\newblock \emph{arXiv preprint arXiv:1704.04368}.

\bibitem[{Shum et~al.(2018)Shum, He, and Li}]{shum2018eliza}
Heung-Yeung Shum, Xiao-dong He, and Di~Li. 2018.
\newblock From eliza to xiaoice: challenges and opportunities with social
  chatbots.
\newblock \emph{Frontiers of Information Technology \&amp; Electronic
  Engineering}, 19(1):10--26.

\bibitem[{Song et~al.(2018)Song, Zhang, Wang, and Gildea}]{song2018graph}
Linfeng Song, Yue Zhang, Zhiguo Wang, and Daniel Gildea. 2018.
\newblock A graph-to-sequence model for amr-to-text generation.
\newblock \emph{arXiv preprint arXiv:1805.02473}.

\bibitem[{Sutskever et~al.(2014)Sutskever, Vinyals, and
  Le}]{sutskever2014sequence}
Ilya Sutskever, Oriol Vinyals, and Quoc~V Le. 2014.
\newblock Sequence to sequence learning with neural networks.
\newblock In \emph{Advances in neural information processing systems}, pages
  3104--3112.

\bibitem[{Vaswani et~al.(2017)Vaswani, Shazeer, Parmar, Uszkoreit, Jones,
  Gomez, Kaiser, and Polosukhin}]{vaswani2017attention}
Ashish Vaswani, Noam Shazeer, Niki Parmar, Jakob Uszkoreit, Llion Jones,
  Aidan~N Gomez, {\L}ukasz Kaiser, and Illia Polosukhin. 2017.
\newblock Attention is all you need.
\newblock In \emph{Advances in Neural Information Processing Systems}, pages
  5998--6008.

\bibitem[{Wang et~al.(2018)Wang, Ye, and Gupta}]{wang2018zero}
Xiaolong Wang, Yufei Ye, and Abhinav Gupta. 2018.
\newblock Zero-shot recognition via semantic embeddings and knowledge graphs.
\newblock In \emph{Proceedings of the IEEE Conference on Computer Vision and
  Pattern Recognition}, pages 6857--6866.

\bibitem[{Xu et~al.(2018{\natexlab{a}})Xu, Wu, Wang, and
  Sheinin}]{xu2018graph2seq}
Kun Xu, Lingfei Wu, Zhiguo Wang, and Vadim Sheinin. 2018{\natexlab{a}}.
\newblock Graph2seq: Graph to sequence learning with attention-based neural
  networks.
\newblock \emph{arXiv preprint arXiv:1804.00823}.

\bibitem[{Xu et~al.(2018{\natexlab{b}})Xu, Wu, Wang, Yu, Chen, and
  Sheinin}]{xu2018sql}
Kun Xu, Lingfei Wu, Zhiguo Wang, Mo~Yu, Liwei Chen, and Vadim Sheinin.
  2018{\natexlab{b}}.
\newblock Sql-to-text generation with graph-to-sequence model.
\newblock \emph{arXiv preprint arXiv:1809.05255}.

\bibitem[{Yang et~al.(2016)Yang, Yang, Dyer, He, Smola, and
  Hovy}]{yang2016hierarchical}
Zichao Yang, Diyi Yang, Chris Dyer, Xiaodong He, Alex Smola, and Eduard Hovy.
  2016.
\newblock Hierarchical attention networks for document classification.
\newblock In \emph{Proceedings of the 2016 Conference of the North American
  Chapter of the Association for Computational Linguistics: Human Language
  Technologies}, pages 1480--1489.

\bibitem[{Yao et~al.(2018)Yao, Mao, and Luo}]{yao2018graph}
Liang Yao, Chengsheng Mao, and Yuan Luo. 2018.
\newblock Graph convolutional networks for text classification.
\newblock \emph{arXiv preprint arXiv:1809.05679}.

\bibitem[{Ying et~al.(2018)Ying, He, Chen, Eksombatchai, Hamilton, and
  Leskovec}]{ying2018graph}
Rex Ying, Ruining He, Kaifeng Chen, Pong Eksombatchai, William~L Hamilton, and
  Jure Leskovec. 2018.
\newblock Graph convolutional neural networks for web-scale recommender
  systems.
\newblock \emph{arXiv preprint arXiv:1806.01973}.

\bibitem[{Zhang et~al.(2018)Zhang, Liu, Niu, Lai, and
  Xu}]{DBLP:conf/cikm/ZhangLNLX18}
Ting Zhang, Bang Liu, Di~Niu, Kunfeng Lai, and Yu~Xu. 2018.
\newblock \href {https://doi.org/10.1145/3269206.3271806} {Multiresolution
  graph attention networks for relevance matching}.
\newblock In \emph{Proceedings of the 27th {ACM} International Conference on
  Information and Knowledge Management, {CIKM} 2018, Torino, Italy, October
  22-26, 2018}, pages 933--942.

\bibitem[{Zhao et~al.(2018)Zhao, Li, Wang, Qian, and Fu}]{zhao2018graphseq2seq}
Guoshuai Zhao, Jun Li, Lu~Wang, Xueming Qian, and Yun Fu. 2018.
\newblock Graphseq2seq: Graph-sequence-to-sequence for neural machine
  translation.

\bibitem[{Zhou et~al.(2018)Zhou, Cui, Zhang, Yang, Liu, and
  Sun}]{DBLP:journals/corr/abs-1812-08434}
Jie Zhou, Ganqu Cui, Zhengyan Zhang, Cheng Yang, Zhiyuan Liu, and Maosong Sun.
  2018.
\newblock \href {http://arxiv.org/abs/1812.08434} {Graph neural networks: {A}
  review of methods and applications}.
\newblock \emph{CoRR}, abs/1812.08434.

\end{thebibliography}
\bibliographystyle{acl_natbib}

\end{document}